\documentclass[letterpaper, 10 pt, conference]{ieeeconf}  % Comment this line out if you need a4paper

\IEEEoverridecommandlockouts                              % This command is only needed if 
\usepackage{multirow}
\usepackage{bbding}
\usepackage{hyperref}
\usepackage{soul}
\usepackage{pifont}
\usepackage[nointegrals]{wasysym}
\usepackage{caption}
\usepackage{subcaption}
\usepackage{amsmath}
\usepackage{array}
\usepackage{color}
\usepackage[normalem]{ulem}
\usepackage{hyperref}
\usepackage{nomencl}
\makenomenclature
\usepackage[ruled,vlined]{algorithm2e}
\usepackage{amsmath,amssymb,amsfonts}
\usepackage{algorithmic}
\usepackage{graphicx}
\usepackage{textcomp}
\usepackage{wrapfig}
\usepackage{mdframed,lipsum}
\usepackage[T1]{fontenc} % I’m accustomed to always loading it...
\usepackage{booktabs}
\usepackage{mwe}
\usepackage[table, dvipsnames]{xcolor}
\usepackage{array,booktabs}

\usepackage{cite}
\usepackage{cleveref}

\usepackage{caption}
\SetKwInput{KwInput}{Input}
\SetKwInput{KwOutput}{Output}

\title{\LARGE \bf
DnD Filter: Differentiable State Estimation for Dynamic Systems using Diffusion Models
}

\author{ Ziyu Wan and Lin Zhao% <-this % stops a space% <-this % stops a space
\thanks{These authors are with the Department of Electrical and Computer Engineering,
        National University of Singapore, 4 Engineering Drive 3, 117583 Singapore, Singapore
        {\tt\small wziyu@u.nus.edu, elezhli@nus.edu.sg}}%
}

\begin{document}

\makeatletter
\let\@oldmaketitle\@maketitle%
\renewcommand{\@maketitle}{\@oldmaketitle%
    \centering
    \vspace*{1mm}
    \includegraphics[width=\textwidth]{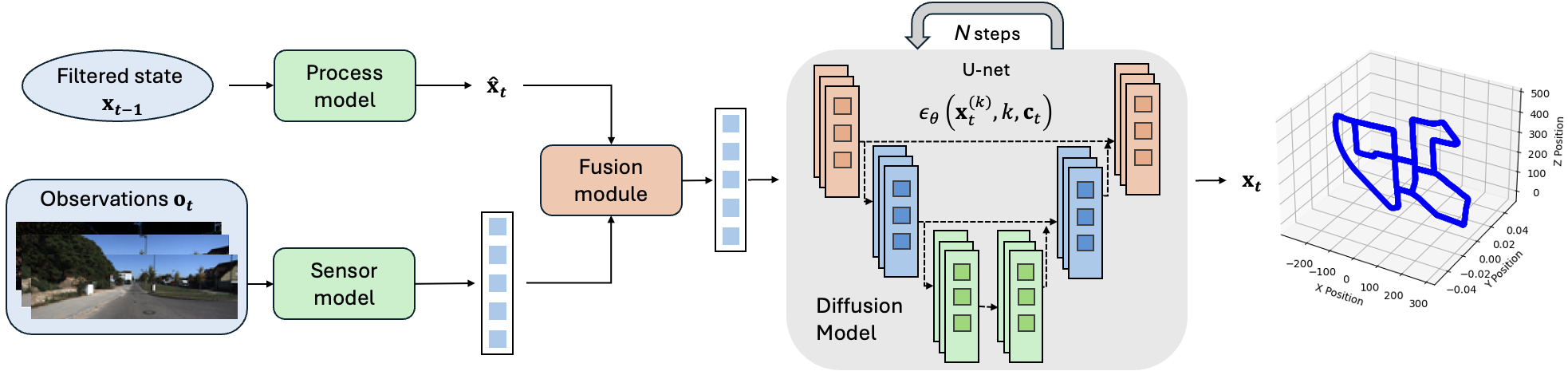}
    \captionof{figure}{An example of employing the DnD Filter on the KITTI visual odometry dataset. DnD Filter is a nonlinear differentiable filter for dynamic system state estimation that utilizes diffusion models. The diffusion model refines the posterior state estimate through a denoising process, conditioned on both the predicted state from the dynamic model and the observed data, while dynamically prioritizing the most accurate sources. Our approach significantly improves state estimation, particularly for high-dimensional and nonlinear scenes, and outperforms state-of-the-art differentiable filters.}
    \label{fig:teaser}
    \vspace*{-3mm}
}
\makeatother

\maketitle
\thispagestyle{empty}
\pagestyle{empty}

%%%%%%%%%%%%%%%%%%%%%%%%%%%%%%%%%%%%%%%%%%%%%%%%%%%%%%%%%%%%%%%%%%%%%%%%%%%%%%%%

\begin{abstract}

This paper proposes the DnD Filter, a \textit{differentiable} filter that utilizes \textit{diffusion} models for state estimation of dynamic systems. Unlike conventional differentiable filters, which often impose restrictive assumptions on process noise (e.g., Gaussianity), DnD Filter enables a nonlinear state update without such constraints by conditioning a diffusion model on both the predicted state and observational data, capitalizing on its ability to approximate complex distributions. We validate its effectiveness on both a simulated task and a real-world visual odometry task, where DnD Filter consistently outperforms existing baselines. Specifically, it achieves a 25\% improvement in estimation accuracy on the visual odometry task compared to state-of-the-art differentiable filters, and even surpasses differentiable smoothers that utilize future measurements. To the best of our knowledge, DnD Filter represents the first successful attempt to leverage diffusion models for state estimation,  offering a flexible and powerful framework for nonlinear estimation under noisy measurements. The code is available at ~\url{https://github.com/ZiyuNUS/DnDFilter}.

\end{abstract}

%%%%%%%%%%%%%%%%%%%%%%%%%%%%%%%%%%%%%%%%%%%%%%%%%%%%%%%%%%%%%%%%%%%%%%%%%%%%%%%%
\section{INTRODUCTION}

Accurate state estimation is fundamental in robotics, enabling autonomous systems to perceive their surroundings, make decisions, and interact with dynamic environments. Traditional approaches, such as Kalman filters (KF)\cite{thrun2005probabilistic}, have been widely used for this purpose \cite{chen2011kalman, wang2022imitation, kaufmann2023champion}, denoising sensor data and estimating system states recursively based on Bayesian theory. However, standard KFs rely on linear Gaussian assumptions, which limit their effectiveness in handling complex, high-dimensional, and nonlinear systems. 
To overcome these limitations, methods like particle filters and ensemble Kalman Filters (EnKFs) \cite{evensen2003ensemble} extend the classical Kalman framework by approximating the posterior state distribution using an ensemble of samples. However, these filters suffer from the curse of dimensionality when dealing with high-dimensional state space. Moreover, the aforementioned canonical filters cannot utilize high-dimensional raw sensing data (e.g., images) for state estimation directly.

In recent years, deep learning approaches have gained traction in state estimation tasks. Deep State-Space Models (DSSMs) \cite{rangapuram2018deep} learn complex system dynamics and measurement models directly from data, overcoming limitations in handling large-scale observations and thus, offering a more flexible solution for nonlinear systems. Differentiable Filters (DFs), a subset of algorithms derived from DSSMs \cite{klushyn2021latent, kloss2021train}, apply the principles of recursive Bayesian estimation to the learned system dynamics and the measurement models \cite{liu2023alpha} for state estimation.
% Due to these characteristics, DFs can adapt more flexibly to real-world scenarios when dealing with systems with intricate dynamics and challenging sensor information like images.
However, most DFs still impose certain assumptions on the characteristics of the process noise, such as Gaussianity. When real noise distribution deviates from these assumptions, the state estimation accuracy  can degrade significantly. In particular, differentiable Kalman filters rely on a Kalman gain computed in a closed-form manner for state updates, which is practically non-differentiable~\cite{liu2023alpha}. Moreover, their updates only linearly combine the observation and prediction, making them overly conservative and leading to poor performance in highly nonlinear dynamic systems.

To mitigate the limitation, we leverage the strong capability of diffusion models to capture complex multi-modal probability distributions, enabling more effective posterior state modeling in the state update within the filtering framework. Specifically, DnD implements a learnable nonlinear state update module using a diffusion model, conditioned on both predicted states and observed data. Figure \ref{fig:teaser} illustrates an example of applying DnD to the KITTI visual odometry dataset~\cite{geiger2013vision}. Compared to Kalman filters, DnD requires no assumptions on noise distributions or covariance for computing the update gain. Moreover, it provides a nonlinear, differentiable update rule, making it suitable for end-to-end learning. Overall, the proposed DnD combines the strengths of diffusion models and differentiable filters to enhance state estimation accuracy. More experimental results are shown in~\Cref{sec:exp}.

The main contributions of this paper can be summarized as follows:
\begin{itemize}
    \item We propose a novel differentiable filtering framework that leverages diffusion models to estimate the posterior state distribution. The state estimate is obtained through the progressive denoising process inherent in diffusion models, eliminating the need for explicit noise modeling or prior assumptions on noise distributions.
    \item To the best of our knowledge, this is the first integration of diffusion models into differentiable filters. By employing diffusion models, our approach adaptively adjusts the reliance on observation and state transition information based on their quality.
    \item We propose a stagewise training approach to accelerate end-to-end training for our DnD Filter, achieving a 20× speedup over traditional iterative training with enhanced training stability.
    \item Empirical evaluations on the KITTI visual odometry dataset show that our method reduces translational and rotational errors by 23\% and 28\%, respectively, compared to state-of-the-art differentiable filters, while also outperforming advanced differentiable smoothers.
\end{itemize}

\section{RELATED WORK}
\subsection{State Estimation}
State estimation aims to infer the hidden states of a dynamic system from noisy observations with models of the system's dynamics and measurement process. Filtering is a key technique in state estimation, used to recursively infer states from observations \cite{thrun2005probabilistic}. Bayesian filtering methods, such as the Kalman filter and its extensions, have been widely applied due to their strong probabilistic foundation \cite{sorenson1985kalman, van2004sigma}. However, traditional KFs assume known system dynamics and Gaussian noise distributions, which limit their applicability in high-dimensional and nonlinear environments.

To address these limitations, sample-based methods such as particle filters approximate the posterior distribution using an ensemble of weighted samples, making them effective for nonlinear and non-Gaussian filtering tasks \cite{gustafsson2010particle}. However, these methods still rely on a large number of particles to maintain accuracy, making them computationally expensive, especially in high-dimensional state spaces.

\subsection{Differentiable Filters}
Recent advancements in deep learning have led to the development of Differentiable Filters, which replace prescribed first-principle models with learned neural network representations \cite{corenflos2021differentiable, piga2021differentiable, li2023differentiable}. DFs learn recursive Bayesian filtering by backpropagation through time, allowing end-to-end training. BackpropKF \cite{haarnoja2016backprop} incorporates convolutional networks into the KF framework, using neural networks to model the measurement process. For sampling-based methods, Differentiable Particle Filter \cite{chen2021differentiable, jonschkowski2018differentiable} incorporates learnable modules for gradient-based optimization. Various improvements, including algorithmic priors and  adversarial techniques \cite{wang2019dualsmc} have been proposed to enhance efficiency and refine learning performance. In addition, DEnKF \cite{liu2023enhancing} incorporates stochastic neural networks into ensemble Kalman filter for differentiable system modeling. 

In \cite{kloss2021train}, extensive experiments on various differentiable filters demonstrate the effectiveness of end-to-end training and provide a comprehensive analysis of their ability to model complex uncertainty distributions. For real-world visual odometry tasks, \cite{haarnoja2016backprop} and \cite{liu2023enhancing} report significant improvements in estimation accuracy. Additionally, differentiable filters have been applied to robotic manipulation tasks, as shown in \cite{lee2020multimodal} and \cite{zachares2021interpreting}, achieving notable performance gains. These results suggest that differentiable filters can effectively capture system dynamics, measurement processes, and noise distributions in practical applications.

While DFs offer increased flexibility and scalability, they still face challenges. Most existing works assume predefined process noise distributions, but deviations in real noise can significantly degrade state estimation accuracy. Furthermore, the Kalman gain structure within the Kalman framework is typically computed in a closed-form manner, limiting full integration into end-to-end learning. Moreover, a linear combination of the prediction and the observation using Kalman gain tends to be conservative. Recent works have attempted to address these issues by introducing attention mechanisms \cite{liu2023alpha}, but these solutions remain constrained when facing intricate multimodal environments.

\begin{figure*}[!h]
    \centering
    \addtocounter{figure}{-1}
    \includegraphics[width=1.0\textwidth]{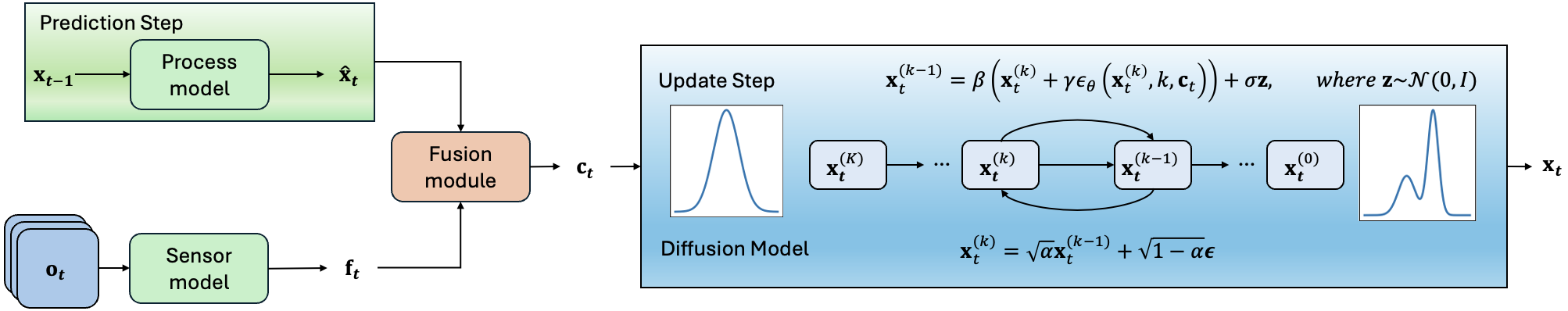}
    \caption{The estimated state at time $t-1$ is input into the process model to predict the state at time $t$, while the observation at time $t$ is processed by the sensor model to extract features. These are then combined by the fusion model, serving as the condition for the diffusion model, which dynamically balances the information sources through its denoising process to estimate the state at time $t$.}
    \label{fig:method}
    \vspace{-0.4cm}  
\end{figure*}

\subsection{Diffusion Models for Probabilistic Modeling}
Diffusion models have recently emerged as powerful frameworks for modeling complex distributions in visual, auditory, and textual data. By iteratively refining noisy inputs through a learned denoising process \cite{sohl2015deep, ho2020denoising}, they become well-suited for tasks with inherent uncertainty and multimodal data distributions. Due to their strong capability in fitting complex distributions and generating conditionally sampled outputs, researchers have explored various approaches to applying diffusion models in reinforcement learning \cite{zhu2023diffusion}, time series forecasting \cite{rasul2021autoregressive}, and decision-making \cite{ajay2022conditional,dong2023aligndiff}. 

In robotics applications, recent studies, such as Diffusion Policy \cite{chi2023diffusion} and NoMaD \cite{sridhar2024nomad}, have shown the effectiveness of conditional diffusion models in generating action sequences by accurately capturing the inherently multi-modal nature of action distributions \cite{wang2022diffusion}. Meanwhile, diffusion models have gained significant attention for their effectiveness in trajectory generation \cite{carvalho2023motion, liang2023adaptdiffuser}. Their strong conditional sampling capability enables generating entire trajectories at once rather than step by step, thereby reducing compounding errors and short-sightedness \cite{dong2024diffuserlite}. Furthermore, a recent study has explored the application of diffusion models in modeling real-world quadrotor dynamics \cite{das2024dronediffusion}.

Despite their success in the above applications, diffusion models remain largely unexplored for state estimation.  Given their strength in modeling complex and multi-modal distributions, diffusion models are ideal in Bayesian estimation to approximate the posterior distribution. This paper presents the first framework that integrates diffusion models with differentiable filtering, which significantly improves the estimation accuracy.

\section{Method}

The objective of a recursive Bayesian differentiable filter is to infer the state $\mathbf{x}_t$  of a dynamic system at time $t$ from a sequence of noisy observations $\mathbf{o}_{1:t}$. This is achieved by iteratively estimating the posterior distribution $p(\mathbf{x}_t | \mathbf{o}_{1:t})$ by incorporating prior knowledge and new observations \cite{sarkka2023bayesian}. The general form of a recursive Bayesian filter consists of two steps:  

1. \textbf{Prediction Step}: The prior distribution of the state at time $t$ is obtained using the system's transition model:
   \begin{equation}
       p(\mathbf{x}_t | \mathbf{o}_{1:t-1}) = \int p(\mathbf{x}_t | \mathbf{x}_{t-1}) p(\mathbf{x}_{t-1} | \mathbf{o}_{1:t-1}) d\mathbf{x}_{t-1}.
   \end{equation}
   This step propagates the belief of the state forward in time based on the system dynamics.
   
2.\textbf{ Update Step}: Given a new observation $\mathbf{o}_t$, the posterior distribution is computed using Bayes' rule:
   \begin{equation}
       p(\mathbf{x}_t | \mathbf{o}_{1:t}) = \frac{p(\mathbf{o}_t | \mathbf{x}_t) p(\mathbf{x}_t | \mathbf{o}_{1:t-1})}{p(\mathbf{o}_t | \mathbf{o}_{1:t-1})}.
   \end{equation}
   This step incorporates new sensor measurements to refine the state estimate.

While Kalman filter assumes linear Gaussian models for state transition and observation, differentiable filters often relax these assumptions by incorporating neural networks to learn transition model $p(\mathbf{x}_t | \mathbf{x}_{t-1})$ and measurement model $p(\mathbf{o}_t | \mathbf{x}_t)$ from data while preserving the recursive Bayesian filtering structure. Our DnD filter is a differentiable recursive Bayesian filter that utilizes the diffusion model to learn the posterior state distributions as in the Update Step. The overall structure of DnD Filter is illustrated in Fig. \ref{fig:method}.

\subsection{Prediction Step and Sensor Model}
Given a system state $\mathbf{x}_{t-1}$ at time $t-1$, the differentiable filter predicts the next state ${\bf \hat{x}}_{t}$ using a transition function:
\begin{equation}
    \begin{aligned}\label{eq:1}
          {\bf \hat{x}}_{t} & \thicksim  g (\ \cdot\ |{\bf x}_{t-1}),\ 
    \end{aligned}
\end{equation}
where $g$ represents the transition dynamics. This step provides an initial estimate of the latent state, which is refined in the subsequent update phase. Various methods have been proposed for modeling state transition dynamics. Traditional approaches are often based on the physical properties of the system or  system identification \cite{ljung1998system}. In recent years, studies have explored the use of Multi-Layer Perceptron (MLP) and stochastic neural networks \cite{jospin2022hands}  to model state transitions. An appropriate process model can be selected based on the specific requirements of the task.

Meanwhile, the high-dimensional observation $\mathbf{o}_t$ obtained at time $t$ (e.g., images) will be fed into a learnable sensor model $\mathbf{s}_\theta$, which extracts an feature representation:
\begin{equation} 
    \begin{aligned}
          {\bf f}_{t} & \thicksim  \mathbf{s_\theta} (\ \cdot\ |{\bf o}_{t-N:t}).\ 
    \end{aligned}
\end{equation}
Unlike conventional differentiable filters that employ neural networks to encode high-dimensional observations and separately extract noise characteristics before the update step, our approach requires only a single encoded feature ${\bf f}_{t}$, eliminating the need for explicitly learning the noise and following the restrictive Kalman filter update rule.

\subsection{Fusion Module and Update Step} \label{subsec:fusion}
A learnable fusion module is introduced to effectively integrate the information from the process model and sensor model:
\begin{equation} 
    \begin{aligned}
          {\bf c}_{t} & =  {\bf Fusion} ({\bf f}_{t}, {\bf \hat{x}}_{t}),\ 
    \end{aligned}
\end{equation}
where ${\bf c}_t$ is the fused feature vector, serving as the conditional input to the diffusion model. This module can be task-specific and implemented with different architectures. The design of the fusion module is flexible: for vector-level fusion, MLP can be employed (see~\Cref{subsec:fusion}), whereas for image-level fusion, it can be integrated into the sensor model (see~\Cref{subsec:disktracking}).

Update step plays a crucial role in adjusting the predicted state by correcting it based on the latest measurement, refining the estimate of the system's state with reduced uncertainty. In KFs, this is typically done by utilizing the relative uncertainty of the prediction and the measurement to calculate the Kalman gain, which requires certain assumptions to explicitly model noise characteristics. Instead of relying on a closed-form parametric update rule, we employ a diffusion model to approximate the posterior distribution $p(\bf{x}_t|o_t, \hat{x}_t)$, allowing for a more flexible update mechanism. 

The state update is formulated as a reverse diffusion process — a stepwise denoising Markov chain that gradually removes noise from pure noise using a learned denoising network, ultimately reconstructing a data sample. Specifically, we implement it using the Denoising Diffusion Probabilistic Model (DDPM), which is known for its ability to effectively capture complex distributions, ensure stable training, and demand minor task-specific parameter adjusting. We start by sampling a random Gaussian noise $\mathbf{x}_t^{(K)}$:
\begin{equation}
   \mathbf{x}_t^{(K)} \sim \mathcal{N}(0, I).
\end{equation}
This noise serves as the initial input, and is gradually refined through an iterative denoising process with a total of $K$ diffusion steps. At each diffusion step $k$, a neural network predicts the noise component $\epsilon_\theta(\mathbf{x}_t^{(k)}, k, {\mathbf c}_t)$, which is then used to estimate the cleaner version of the perturbed state at the previous timestep $k-1$:
% \begin{equation}
% \begin{aligned}
% \bf{x}_{t}^{(k-1)}  = &\frac{1}{\sqrt{\alpha}} \left( \bf{x}_{t}^{(k)} - \frac{1 - \alpha}{\sqrt{1 - \bar{\alpha}}} \epsilon_{\theta}(x_{t}^{(k)}, k, {\bf c}_t) \right) + \sigma \bf{z}, \\
% := &\beta(x_t^{(k)}+\gamma\epsilon_\theta(x_t^{(k)},k,{\bf c}_t))+ \sigma z,\\
% \quad \text{where} \quad {z} \sim \mathcal{N}(0, {I}) &
% \end{aligned}
% \end{equation}
\begin{equation}
\begin{aligned}
\mathbf{x}_{t}^{(k-1)} &= \frac{1}{\sqrt{\alpha}} \left( \mathbf{x}_{t}^{(k)} - \frac{1 - \alpha}{\sqrt{1 - \bar{\alpha}}} \epsilon_{\theta}(\mathbf{x}_{t}^{(k)}, k, \mathbf{c}_t) \right) + \sigma \mathbf{z}, \\
&:= \beta(\mathbf{x}_t^{(k)} + \gamma\epsilon_\theta(\mathbf{x}_t^{(k)}, k, \mathbf{c}_t)) + \sigma \mathbf{z},
\end{aligned}
\end{equation}
% \begin{equation}
% \begin{aligned}
%     x_{t}^{(k-1)} &= \frac{1}{\sqrt{\alpha}} \left( x_{t}^{(k)} - 
%     \frac{1 - \alpha}{\sqrt{1 - \bar{\alpha}}} \epsilon_{\theta}(x_{t}^{(k)}, k, c_t) \right) 
%     + \sigma z, \\
%     &\hfill \quad \quad \quad \quad \quad \quad \quad \quad \quad \quad \quad \quad \quad \quad z \sim \mathcal{N}(0, I)\\
%     &= \beta(x_t^{(k)}+\gamma\epsilon_\theta(x_t^{(k)},k,c_t))+ \mathcal{N}(0,\sigma^2I)
% \end{aligned}
% \end{equation}
where $\mathbf{z} \sim \mathcal{N}(0, I)$, $\mathbf{x}_{t}^{(k)}$ denotes the perturbed state at the $k$-th diffusion step, \( \alpha \), \( \bar{\alpha} \), \( {\sigma} \) are predefined noise scaling parameters~\cite{chan2024tutorial}, $\beta $ and $\gamma$ denote the rescaling factor introduced for short notation. This reverse diffusion process progressively removes noise, iterating from $K$ to $0$, reconstructing a sample that follows the learned data distribution. The final output $\mathbf{x}_t^{(0)}$ is obtained and serves as the filtered state $\mathbf{x}_t$. This process enables a flexible nonlinear combination between the observation and the prediction. Due to the stochastic denoising process inherent in diffusion models, the diffusion model-embedded update step enables a adaptable trade-off between the observation and the prediction. 

\subsection{End-to-End Learning}
 % Traditional differentiable filters are typically trained end-to-end using the mean squared error (MSE) between the intermediate or final outputs of the model and the ground truth as the loss function. However, with the incorporation of the diffusion module, our end-to-end training must be conducted within the training framework of diffusion model.
The training process of a DDPM consists of two main steps: a forward diffusion process and the learning of a noise prediction model. In the forward diffusion process, noise is gradually added to a ground truth state \(\mathbf{x}_t^* = \mathbf{x}^{(0)}_t \) over \( K \) timesteps until it becomes nearly Gaussian noise. During training,  the entire sequential noising process is not simulated; instead, a noisy sample at any timestep  \( k \) is directly computed using the closed-form equation:
\begin{equation}
\begin{aligned}
    \mathbf{x}_t^{(k)} = \sqrt{\bar{\alpha}} \mathbf{x}_t^{(0)} + \sqrt{1 - \bar{\alpha}} \epsilon, \quad \epsilon \sim \mathcal{N}(0, I),
\end{aligned}
\end{equation}
where $\epsilon$ denotes the added noise.
The goal of training is to learn a neural network $\epsilon_\theta(\mathbf{x}_t^{(k)}, k, {\bf c}_t)$ to predict the noise component $\epsilon$, given a noisy sample \( \mathbf{x}_t^{(k)} \), its corresponding timestep \( k \) and the condition ${\bf c}_t$. We use a U-Net as the noise prediction network (see~\Cref{fig:teaser}). The model $\epsilon_\theta(\mathbf{x}_t^{(k)}, k, {\bf c}_t)$ is trained by minimizing the loss between the predicted and actual noise:
\begin{equation}
\begin{aligned}
\mathcal{L} = \mathbb{E}_{\mathbf{x}_t^{(k)}, k, {\bf c}_t, \epsilon} \left[ \left\| \epsilon - \epsilon_\theta(\mathbf{x}_t^{(k)}, k, {\bf c}_t) \right\|^2 \right].
\end{aligned}
\end{equation}
During training, the following steps are repeated:
(1) Sample a ground truth state \(\mathbf{x}_t^* = \mathbf{x}^{(0)}_t \). (2) Generate the noisy sample \( \mathbf{x}_t^{(k)} \) using the closed-form equation. (3) Predict the noise using the neural network \( \epsilon_\theta(\mathbf{x}_t^{(k)}, k, {\bf c}_t) \) to predict the noise \( \epsilon \). 
(4) Compute and minimize the MSE loss. The process model, sensor model, and fusion module will all be jointly trained with diffusion model to ensure that the hyper-parameters are optimized consistently.
\begin{figure}[!tpb]
  \centering
  \vspace{0.2cm}
  \includegraphics[scale = 0.48]{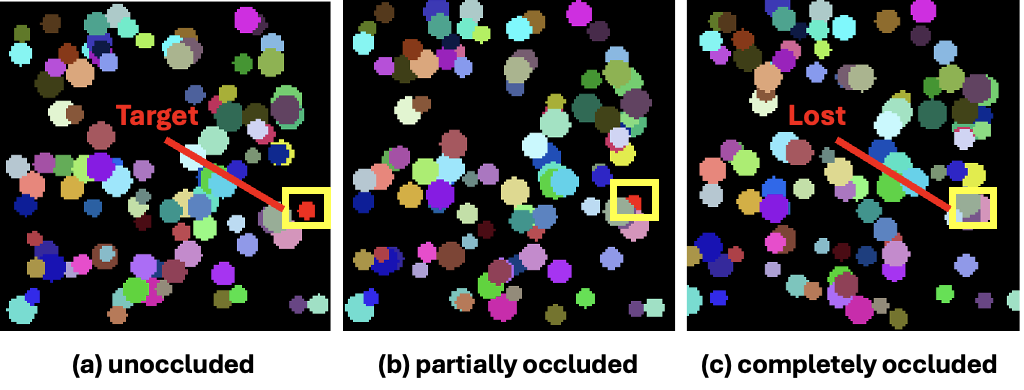}
  \vspace{-0.35cm}
  \caption{Three images sampled from a sequence, where the yellow box highlights the target disk. (a) shows the unobstructed case, where the observation uncertainty is minimal. (b) shows the partially occluded case, where the observation uncertainty increases. (c) shows the fully occluded case, where the observation uncertainty is at its highest.}
  \label{fig:disk_example}
  \vspace{-0.7cm}  
\end{figure}
\section{EXPERIMENTS} \label{sec:exp}
We test our DnD filter on a simulated experiment and on a real-world dataset to verify its effectiveness. 
(1) Can the DnD filter be generally applied across different scenarios?
(2) Compared to other state-of-the-art differentiable filters, how much does the introduction of the diffusion model improve the accuracy of state estimation?
(3) Considering the iterative denoising process of the diffusion mechanism, can the DnD filter maintain a satisfactory inference frequency to ensure practical usability?

\subsection{Vision-based Disk Tracking} \label{subsec:disktracking}
\textbf{Overview:} The task is to keep tracking of the trajectory of the target disk from a sequence of observed images in a 2D scene with numerous interfering disks. This task becomes challenging as the number of interfering disks grows. For filtering algorithms, it requires maintaining tracking when the target is occluded, handling noise from various sources, and processing high-dimensional inputs. In our
experiments, the number of the interfering disks is set to 100, indicating that the target disk may be occluded most of the time. A example of the observed images is shown in Fig. \ref{fig:disk_example}. We define the position of the target disk as the state.

\textbf{Data:} We split the dataset into a training set and a test set, with the training set containing 500 image sequences and the test set containing 100 image sequences. Each image sequence consists of 100 128×128 RGB images.
The target disk has a fixed color but a random size, while the interfering disks have both randomly generated colors and sizes. The interfering disks may leave the image frame. The motion of all disks is governed by a linear-Gaussian transition law:
\begin{equation}
\begin{aligned}
    \left\{
    \begin{array}{ll}
        \mathbf{x}^*_{t+1} = \mathbf{x}^*_{t} + \mathbf{v} + 1.5\times \omega_{t} \\
        \mathbf{v} = \text{Const.} \\
        \omega_{t} \sim \mathcal{N}(0,1),
    \end{array}
    \right.
\end{aligned}
\label{eq:motion}
\end{equation}
where $\mathbf{x}^*_t$ and $\mathbf{v}$ denote the ground truth position and velocity of the target disk, respectively, with randomly sampled initial values.

\textbf{Filter Implementation:} We utilize the following transition model (process model):
\begin{equation}
\begin{aligned}
    \left\{
    \begin{array}{ll}
        \hat{\mathbf{x}}_{t+1} = \mathbf{x}_{t} + \mathbf{v}\\
        \mathbf{v} = \text{Const.} \\
    \end{array}
    \right.
\end{aligned}
\end{equation}
The noise in the simulated disk movement accumulates over time, leading to drift from the predicted states generated by the process model. To mitigate this, we incorporate observations to refine the predictions. 
Specifically, we select images observed at $t$, $t-2$, $t-4$ and $t-6$ as observations $\mathbf{o}_t$. A VinT  \cite{shah2023vint} is used to construct the sensor model. To effectively fuse the predicted state $\hat{\mathbf{x}}_t$ with observation $\mathbf{o}_t$, we first represent $\hat{\mathbf{x}}_t$ as a heat map. This heatmap is then concatenated with the four images of $\mathbf{o}_t$.  The combined input is fed into the VinT sensor model for feature extraction and fusion. The resulting output features, denoted as  ${\bf c}_t$ , serve as the conditioning input for the diffusion model. A 3-layer U-Net is employed to predict the noise in the diffusion process.

\begin{figure}[!tpb]
  \centering
  % \vspace{-0.2cm}
  \includegraphics[scale = 0.465]{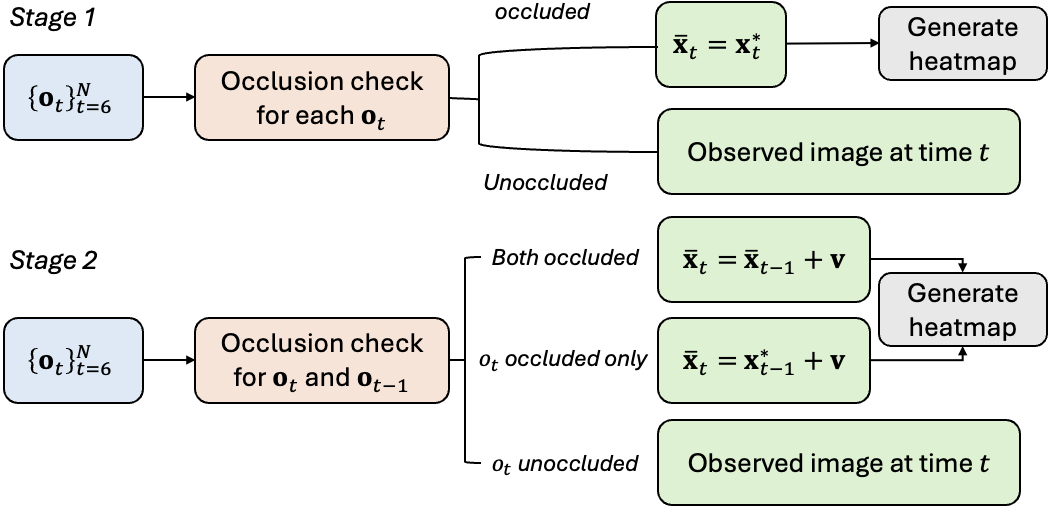}
  \vspace{-0.2cm}
  \caption{The proposed 2-stage training method with progressive ground truth state guidance.}
  \label{fig:disk_training}
  \vspace{-0.45cm}  
\end{figure}

\textbf{Training:} Since the diffusion model requires $K$ denoising steps to generate the estimated state ${\mathbf{x}}_t$ from the white noise, training under the RNN framework (backpropagation through time), commonly used by traditional differentiable filters, would be extremely time-consuming. Inspired by teacher forcing \cite{bengio2015scheduled}, we define $\bar{\mathbf{x}}_t$ as an approximation of $\hat{\mathbf{x}}_t$ and propose a \textbf{stagewise training} method (illustrated in~\Cref{fig:disk_training}), which not only improves the training efficiency, but also progressively guides the DnD model to prioritize the use of different information sources based on their quality. More specifically, 
% The method substitute $\{{x}_t\}_{t=6}^N$ with more accessible $\{\bar{x}_t\}_{t=6}^N$.
\begin{itemize}
    \item In Stage 1, we first determine whether the target is occluded at time $t$. If it is occluded, we set $\bar{\mathbf{x}}_t = \mathbf{x}^*_t$ and represent it using a heatmap; otherwise, we directly use the observed image at time $t$. The resulting representation is then concatenated with $\mathbf{o}_t$ and processed as described in Filter Implementation section. This stage encourages the model to extract useful information from the observations while relying more on $\hat{\mathbf{x}}_t$ when observations become unreliable due to occlusion.

    \item In Stage 2, we follow the same procedure as in the Stage 1 but approximate $\hat{\mathbf{x}}_t$ in a way that better aligns with the inference process when the target is occluded at time $t$:
    \begin{equation}
        \bar{\mathbf{x}}_t =
        \left\{
        \begin{aligned}
            &  \bar{\mathbf{x}}_{t-1} + \mathbf{v} && \text{if target is occluded at $t-1$} \\
            &  {\mathbf{x}}^*_{t-1} + \mathbf{v} && \text{otherwise,} 
        \end{aligned}
        \right.
    \end{equation}
    This stage improves the representation of $\bar{\mathbf{x}}_t$, ensuring greater consistency between training and inference.

    \item Moreover, an optional training stage 3 can be added, which performs iterative end-to-end training under the RNN framework.
\end{itemize}
The key point of our training method is to approximate $\hat{\mathbf{x}}_t$ so that the iterative denoising process can be avoided during training. While Stage 3 training can further improve the estimation performance, 2-stage training is sufficient for most tasks. More importantly, the proposed stagewise training enables parallel processing of the overall model which significantly accelerates training. Compared to straightforward end-to-end training, our approach achieves a 20$\times$ speedup while simultaneously enhances the training stability.

\textbf{Results:} We deployed the trained model in a series of experiments within the iterative RNN architecture, without requiring occlusion indicators as needed during training, in order to demonstrate the following:

1. The superior performance of the DnD Filter compared to other baselines: piecewise Kalman filter, Backprop KF \cite{haarnoja2016backprop}, LSTM, and Transformer; All baselines share the same sensor model and process model structure (if applicable).

2. The inference speed of the DnD filter under different diffusion step settings;

3. Ablation experiments including: 
\begin{itemize}
    \item Effect of diffusion steps: we set the diffusion steps to 5 and 10, resulting in the models DnD(5s) and DnD(10s) after a 3-stage training;
    \item Effect of incorporating the predicted state from the process model: we trained a model that relies solely on observations for information with 10 diffusion steps, DnD(no pred);
    \item Effect of the accuracy of the process model: We evaluate this by replacing the predicted state with the ground truth state in DnD(10s), referred to as DnD(10s with gt);
    \item Effect of our designed training method: We tested DnD(ts1) and DnD(ts2), trained with 1 and 2 stages, respectively, using 10 diffusion steps. Since directly performing iterative end-to-end training under the RNN architecture is highly unstable, it is not included in the comparison;
    \item Effect of introducing the diffusion model: we replaced the diffusion model with an MLP layer, using VinT, the sensor model only, to predict the target position based on both observations and predictions, as well as using observations only. This resulted in models VinT and VinT(wo. pred), trained under the same 2-stage training method.
\end{itemize}
We use the MSE between the estimated trajectory of the target and the ground truth trajectory as the evaluation metric. Experimental results are presented in Table \ref{tab:disk_inference} and Table \ref{tab:disk_performance}. All experiments were conducted on an NVIDIA GeForce RTX 4080 GPU and an Intel i7-14700KF CPU.
\begin{table}[h]
\caption{Inference frequency for different diffusion step settings.}
\centering
\scalebox{1}{
\begin{tabular}{c c c}
    \toprule
    Diffusion Step & T = 5 & T = 10\\ 
    \midrule
    Inference Frequency (Hz)  & 55          & 40           \\
    \bottomrule
\end{tabular}}
\label{tab:disk_inference}
\end{table}

\begin{table}[h]
\centering
\vspace{-0.2cm}
\caption{Ablation studies and benchmarking against state-of-the-art algorithms}
\scalebox{1}{
\begin{tabular}{@{}lll@{}}  
    \toprule
    \multicolumn{1}{c}{Setting} & \multicolumn{1}{c}{Model} & \multicolumn{1}{c}{MSE (pixels)} \\
    \midrule
    \multirow{2}{*}{Diffusion Step} 
    & DnD (5s) & 0.715 $\pm$ 0.048 \\
    & DnD (10s) & \textbf{0.678 $\pm$ 0.050} \\
    \midrule
    \multirow{2}{*}{Predicted State} 
    & DnD (no pred) & 0.855 $\pm$ 0.064 \\
    & DnD (10s with gt) & \textbf{0.515} $\pm$ \textbf{0.039} \\
    \midrule
    \multirow{2}{*}{Training Method} 
    & DnD (ts1) & 1.407 $\pm$ 0.072 \\
    & DnD (ts2) & 0.712 $\pm$ 0.051 \\
    \midrule
    \multirow{2}{*}{Diffusion Model} 
    & VinT & 1.452 $\pm$ 0.090 \\
    & VinT (w.o. pred) & 1.543 $\pm$ 0.122 \\
    \midrule
    \multirow{4}{*}{Prior Works} 
    & Piecewise KF & 1.483 $\pm$ 0.102 \\
    & Backprop KF & 1.217 $\pm$ 0.094 \\
    & LSTM & 1.028 $\pm$ 0.079 \\
    & Transformer & 1.077 $\pm$ 0.077 \\
    \bottomrule
\end{tabular}}
\label{tab:disk_performance}
\vspace{-0.2cm}  
\end{table}

\textbf{Comparison:} The results listed in Table~\ref{tab:disk_inference} and Table~\ref{tab:disk_performance} demonstrate that:

1. DnD demonstrates superior performance compared to prior works, including Piecewise KF, Backprop KF, LSTM, and Transformer, achieving more robust and accurate estimation especially in dynamic scenarios.

2. For $T = 5$ , the inference frequency is 55 Hz, while for $T = 10$, the inference frequency  is 40 Hz, the relatively high inference frequency ensures practical usability.

3. Increasing the diffusion steps from 5 to 10 improves the model's accuracy, with DnD(10s) achieving a lower MSE (0.675 ± 0.499). However, increasing the diffusion steps also leads to a lower inference frequency.

\begin{figure}[h]
  \centering
  \vspace{-0.1cm}
  \includegraphics[scale = 0.57]{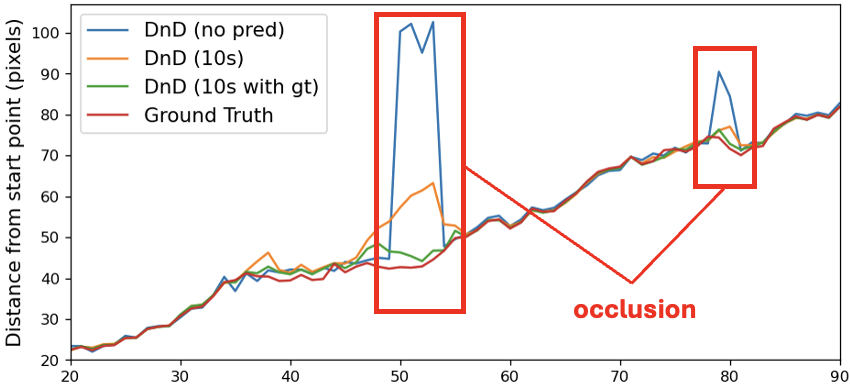}
  \vspace{-0.2cm}
  \caption{The deviation from the ground truth is the largest in the absence of predicted states (blue), while using the ground truth process model (green) for the prediction step yields the closest results to the ground truth.}
  \label{fig:disk_result}
  \vspace{-0.4cm}  
\end{figure}

4. Incorporating the predicted state from the process model significantly improves performance. The more accurate the process model is, the more accurate the overall estimation is. It can be seen in Fig. \ref{fig:disk_result} that the proposed DnD filter can dynamically prioritizing the usage of observation and prediction based on the uncertainty (target occlusion) for estimation.

5. Through our proposed stagewise training method, the model performance improves progressively. Notably, DnD (ts2) with two-stage training achieves performance close to DnD (10s) with three-stage training.

6. Introducing the diffusion model significantly improved performance, demonstrating that using diffusion to approximate the posterior distribution for the update step allows for more effective integration of information from observations and predictions.

\subsection{KITTI Visual Odometry}
\textbf{Overview:} 
This task is based on the KITTI Visual Odometry dataset \cite{geiger2013vision}. Our goal is to estimate the location and heading of a vehicle during its motion using RGB image sequences captured by two cameras and the known initial pose. The challenge of this task is that the vehicle’s pose cannot be directly obtained from the observed images. It requires predicting relative displacement and rotation between consecutive frames and integrating them to get the absolute pose, which can lead to error accumulation due to the lack of loop closure. Another challenge is training a model on a small dataset that generalizes well to unseen road scenarios.

\textbf{Data:} We used the KITTI-10 dataset mentioned in \cite{kloss2021train} to evaluate our method. The dataset contains 10 trajectories, each consisting of image sequences captured from two cameras while a vehicle drove through urban regions, along with ground truth location and heading data.  The data was sampled at approximately 10 Hz. Follow this work, we used images from both cameras along with their mirrored versions for data augmentation. These augmented trajectories were segmented into shorter sequences for training and evaluating, with each frame downsampled to a 50×150 RGB image.

\begin{figure}[h]
  \centering
  \vspace{-0.1cm}
  \includegraphics[scale = 0.137]{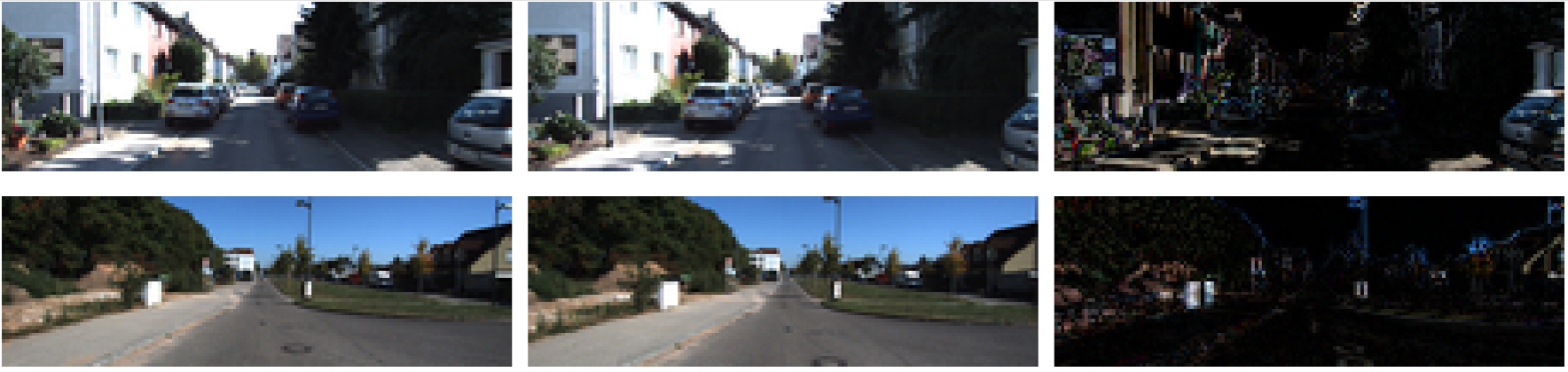}
  \vspace{-0.3cm}
  \caption{Two examples from different trajectories: (Left) Image at $t-1$, (Center) Image at $t$, (Right) Difference image.}
  \label{fig:kitti_example}
  \vspace{-0.2cm}  
\end{figure}

\textbf{Filter Implementation:} We represent the state as a 5-dimensional vector $\mathbf{x}_t = (p^x_t,p^y_t,\theta_t, v_t, \omega_t)$, the same as in \cite{kloss2021train,yi2021differentiable}, where $p^x_t$ and $p^y_t$ represent the position of the vehicle, $\theta_t$ denotes the heading angle, $v_t$ is the linear velocity in the direction of motion, and $\omega_t$ is the angular velocity. For the observation, instead of using images from the current and previous time steps, we use the current image along with a difference image obtained by subtracting the previous image from the current one following~\cite{haarnoja2016backprop} as illustrated in Fig.~\ref{fig:kitti_example}.
Since the dataset does not provide control inputs, we model the variations in $v_t$ and $\omega_t$ using Gaussian noises and formulate the process model as a simple linear model of the translational and angular velocities. We use the same convolutional neural network (CNN) architecture as in \cite{kloss2021train, yi2021differentiable} for the sensor model. For the fusion model, we concatenate the predicted state $\hat{\mathbf{x}}_t$ from the process model with the intermediate feature vector ${\bf f}_t$ from the sensor model to obtain the condition ${\bf c}_t$. 
A U-Net will be used as the noise prediction network for the diffusion model.

\textbf{Training:} We adopt a 10-fold training strategy, using 9 trajectories for training and one trajectory for testing in each iteration. As described in the previous section, we employ a teacher-forcing based 2-stage training approach to accelerate training. 
\begin{itemize}
    \item In Stage 1, we set the predicted state to zero and predict pose changes solely from observed images, serving as a pretraining phase.

    \item In Stage 2, we approximate $\hat{\mathbf{x}}_t$ by either the ground truth state or zero, based on the outcome of a randomly sampled Bernoulli distribution, ensuring equal probability for each choice. This encourages the model to focus more on the predicted state while preventing over-reliance on it.
\end{itemize}

\begin{table}[h]
\centering
\vspace{-0.2cm}
\caption{Performance comparison of DnD and baseline models on KITTI dataset.}
\scalebox{1}{
\begin{tabular}{@{}llll@{}}  
    \toprule
    \multicolumn{1}{c}{Setting} & \multicolumn{1}{c}{Methods} & 
    \multicolumn{1}{c}{m/m} & \multicolumn{1}{c}{deg/m)}\\
    \midrule
    \multirow{2}{*}{U-net Layer} 
    & DnD (3L)  & \textbf{0.130 $\pm$ 0.011}& \textbf{0.0534 $\pm$ 0.005}\\
    & DnD (2L)  & 0.142 $\pm$ 0.016& 0.0554 $\pm$ 0.005\\
    \midrule
    \multirow{4}{*}{\shortstack[l]{Differentiable \\ Filter \cite{kloss2021train,yi2021differentiable}}} 
    & dUKF   & 0.180 $\pm$ 0.023 & 0.0799 $\pm$ 0.008 \\
    & dEKF  & 0.168 $\pm$ 0.012 & 0.0743 $\pm$ 0.007 \\
    & dPF-M-lrn  & 0.190 $\pm$ 0.030 & 0.0900 $\pm$ 0.014\\
    & dMCUKF  & 0.200 $\pm$ 0.030 & 0.0820 $\pm$ 0.013 \\
    \midrule
    \multirow{2}{*}{LSTM} 
    & LSTM (uni.)  & 0.757 $\pm$ 0.012 & 0.3779 $\pm$ 0.054\\
    & LSTM (dy.)  & 0.538 $\pm$ 0.057 & 0.0788 $\pm$ 0.008\\
    \midrule
    \multirow{2}{*}{\shortstack[l]{Smoother \cite{yi2021differentiable}, \\ Heteroscedastic}} 
    & E2E Loss, Vel & 0.148 $\pm$ 0.009& 0.0720 $\pm$ 0.006\\
    & E2E Loss, Pos & 0.146 $\pm$ 0.010& 0.0762 $\pm$ 0.007\\
    \bottomrule
\end{tabular}}
\label{tab:kitti_performance}
\vspace{-0.2cm}  
\end{table}

\textbf{Results:} We set the diffusion step to 10 and adopt the standard KITTI metrics. Specifically, we evaluate our method on 100-frame sequences, compare the final estimated state with the ground truth, and normalize the errors by the total traveled distance, including positional error (m/m) and angular error (deg/m). We compare our results with three groups of baseline methods: 

    1) Differentiable Filters Methods: We utilize state-of-the-art differentiable recursive filters explored in prior works \cite{kloss2021train, haarnoja2016backprop} as baselines. For fairness, our method shares the same input and sensor model network architecture as these baseline methods.

    2) LSTM baseline: Building on previous research in differentiable filtering \cite{kloss2021train, haarnoja2016backprop}, we compare our approach with long short-term memory (LSTM) networks \cite{graves2012long}. We choose an unidirectional LSTM and a LSTM integrated with dynamics as the baselines~\cite{kloss2021train}. 

    3) A smoother based on factor graph optimization:  We use the differentiable smoothers proposed in~\cite{yi2021differentiable} as the baseline. They are based on factor graph optimization. These methods unroll the optimizer for maximum a posteriori inference, enabling learning of probabilistic system models. Since smoothers have access to the entire time sequence, they are theoretically more advantageous than filters.

Also, we investigate the impact of reducing the size of the U-Net employed in the DnD filter. We compare the model performance between a three-layer U-Net of channel sizes $[64, 128, 256]$ with 3.6M parameters and a two-layer U-Net of channel sizes $[64, 128]$ of 1.3M parameters. The latter has fewer parameters than the CNN-based sensor model. All of the experimental results are presented in Table \ref{tab:kitti_performance}.

\textbf{Comparison:} Our proposed diffusion differentiable filter reduces translational and rotational errors by 23\% and 28\%, respectively, compared to other state-of-the-art differentiable filters, while also outperforming advanced differentiable smoothers. This demonstrates how the diffusion model, with its strong ability to fit complex probabilistic distributions, enables the filter to better trade off between observation and prediction when estimating the posterior distribution. Moreover, reducing the number of layers in the noise prediction network of the diffusion model does not lead to a significant performance drop, maintaining relatively good performance.The inference frequency of our method is approximately 45 Hz, which is more than sufficient given the observation sampling rate of 10 Hz.

\section{CONCLUSIONS}
In conclusion, we have presented a novel framework that integrates diffusion models with differentiable filtering to enhance state estimation in dynamic systems. By leveraging the powerful ability of diffusion models to capture complex, multi-modal distributions, our approach provides significant improvements in state estimation within a recursive Bayesian filtering framework. Unlike traditional differentiable filters that rely on assumptions about noise characteristics for state updates, our method introduces a learned denoising process, offering more flexible and probabilistic state updates.

Through extensive experiments on both simulated and real-world datasets, we have demonstrated the superior performance of our approach, achieving substantial improvements over conventional methods. Our model sets a new benchmark in state estimation accuracy.

Our future works include further refining the model to be more lightweight and computationally efficient for real-time applications. Additionally, we plan to incorporate loop closure techniques to correct accumulated errors and explore the integration of additional sensory modalities. Ultimately, our approach paves the way for enhancing recursive Bayesian state estimation using diffusion models, with promising applications in autonomous robots.


%%%%%%%%%%%%%%%%%%%%%%%%%%%%%%%%%%%%%%%%%%%%%%%%%%%%%%%%%%%%%%%%%%%%%%%%%%%%%%%%
%%%%%%%%%%%%%%%%%%%%%%%%%%%%%%%%%%%%%%%%%%%%%%%%%%%%%%%%%%%%%%%%%%%%%%%%%%%%%%%%
%%%%%%%%%%%%%%%%%%%%%%%%%%%%%%%%%%%%%%%%%%%%%%%%%%%%%%%%%%%%%%%%%%%%%%%%%%%%%%%%

\bibliographystyle{IEEEtran}
\bibliography{main}

\end{document}